\documentclass{article}

\usepackage{arxiv}

\usepackage[utf8]{inputenc} 
\usepackage[T1]{fontenc}    
\usepackage{url}            
\usepackage{booktabs}       
\usepackage{amsfonts}       
\usepackage{nicefrac}       
\usepackage{microtype}      
\usepackage{epsfig}
\usepackage{graphicx}
\usepackage{amsmath}
\usepackage{amssymb}
\usepackage{multicol}
\usepackage{url}
\usepackage{algorithm}
\usepackage[noend]{algpseudocode}
\usepackage{subfig}
\usepackage{wrapfig}
\usepackage{color}

\title{Hierarchical Neural Architecture Search via Operator Clustering}
\author{Li Guilin\textsuperscript{1},  Zhang Xing\textsuperscript{1}, Jiashi	Feng\textsuperscript{2},  Wang Zitong\textsuperscript{1}, Matthias	Tan\textsuperscript{4}, Li Zhenguo\textsuperscript{1}, Zhang Tong\textsuperscript{5} \\
\textsuperscript{1}Huawei Noah’s Ark Lab,
\textsuperscript{1}NUS,
\textsuperscript{4}CityU,
\textsuperscript{5}HKUST\\
\{ hiliguilin@gmail.com, zhang.xing1@huawei.com,   tongzhang@tongzhang-ml.org \}}

\begin{document}

\maketitle

\begin{abstract}
Recently, the efficiency of automatic neural architecture design has been significantly improved by gradient-based search methods such as DARTS. However, recent literature has brought doubt to the generalization ability  of DARTS, arguing that DARTS performs poorly when the  search space is changed, i.e, when different set of  candidate operators are used.   Regularization techniques such as early stopping have been proposed to partially solve this problem.  In this paper, we tackle this problem from a different perspective by identifying two  contributing  factors  to the collapse of DARTS when the search space changes: (1) the correlation of similar operators incurs unfavorable  competition among them and makes their relative importance score unreliable and (2) the optimization complexity gap between the proxy search stage and the final training. Based on these findings, we propose a new hierarchical search  algorithm. With its operator clustering and optimization complexity match, the algorithm can consistently find high-performance architecture across various search spaces.  For all the five variants of the popular cell-based search spaces, the proposed algorithm always obtains state-of-the-art architecture with best accuracy on the CIFAR-10, CIFAR-100 and ImageNet over other well-established DARTS-alike algorithms. Code is available at https://github.com/susan0199/StacNAS.
\end{abstract}
\section{Introduction}


Recent years have witnesses the rapid development of automatic neural architecture design since introduction of NAS by Zoph et al \cite{zoph2016neural}.  
To further improve search efficiency, recent work has explored various one-shot architecture search methods \cite{bender2018understanding, liu2018darts, guo2019single,xie2018snas, cai2018proxylessnas,yang2019cars}.
by using  supernet and weight sharing to reduce the cost of training. Among those models, DARTS \cite{liu2018darts} relaxes the discrete search space to be continuous so that stochastic gradient descent can be applied to simultaneously learn the architectures and model parameters.  

However, some recent works have demonstrated one important search collapse problem of DARTS, i.e., the algorithm tends to converge to a cell with all skip-connections, giving under-performing architectures.  To address this problem,  Arber et.al \cite{zela2019understanding}  proposed several regularization techniques to find optima with better generalization property and reported  improved results across several variants of DARTS search spaces.  However, in our experiments, we found that these regularization techniques do not work well on every search space we experimented with, driving us to  further improve their results on the generalization ability of DARTS  from different perspectives. We identify two core issues that lead to the above failure of DARTS-alike algorithms, i.e., the correlation among operators in the supernet; and the optimization complexity gap between the proxy search stage and final training stage. 

\begin{wrapfigure}[24]{r}{0.52\textwidth}
\centering
        \includegraphics[width=0.32\textheight]{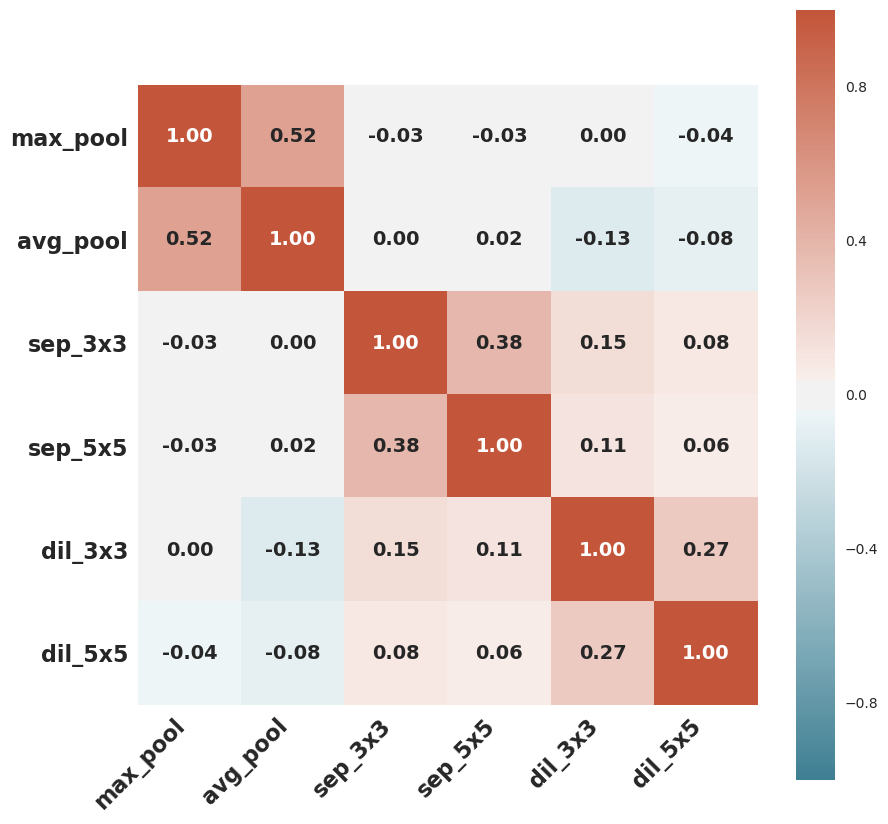}
\caption{Operator correlation matrix: pairwise correlation between flattened feature map of candidate operators. The details of the correlation calculation procedure are provided in the Appendix \ref{Appsec:corr}. } 
\label{fig:corrmatrix}
\end{wrapfigure}

More concretely, the first factor is the \textit{correlation of similar operators} such as \textit{SepConv $3\times3$} and \textit{SepConv $5\times5$}, as illustrated in Figure \ref{fig:corrmatrix}. To calculate this correlation matrix, we first flatten the output feature map tensor into vectors of each operator at a fixed node, and then calculate the  Pearson correlation coefficient between any two of the candidate operators.  When some operators producing highly correlated feature maps, their contribution to the network search should be considered jointly. Otherwise, the architecture parameter cannot be estimated accurately because their contribution score would be decreased due to such unfavorable within-group competition.

The second factor that affects the search performance dramatically is \textit{the gap between the optimization complexity of the proxy search stage   and the final training stage}. 
It is known that skip-connections are crucial to the training of deep CNN models even though they may not be so 
 important for the shallower ones. In DARTS, the author first uses a shallow proxy network with eight-cells to   search for the optimal cell structures, and then repeats the discovered cells to form a much deeper 20-cells  final structure.  
This gap causes the search algorithm to select more than enough complicated operators such as convolution. We empirically find when the parent architecture in the proxy search space is too shallow, the algorithm would tend to select fewer skip connections; however when the parent architecture is too deep, the algorithm would select more than the optimal number of skip connections. By using   \textbf{gradient confusion} \cite{sankararaman2019impact} to measure and explain the  optimization complexity gap problem, we show that a proper depth must be selected if a smaller  supernet is used in the proxy   search stage. 
This complexity matching is essential for the algorithm to  select an appropriate number of skip-connections for the target networks with different depths, while the vanilla DARTS fails to do so.

To solve these problems, we introduce a new hierarchical search algorithm that performs operator clustering and optimization complexity matching.    In the first stage of the algorithm, the best group is activated, and in the second stage, the best operator in the activated group is selected. This hierarchical search algorithm would select proper operators with matching complexities more accurately, and yield a more STAble and Consistent differentiable Neural Architecture Search method, we name this method as \textbf{StacNAS}.     

By solving these two problem, our proposed algorithm can generalize well to different variants of the cell-based search space, where the other existing DARTS-alike algorithms yield degenerate or unstable results. 
In our experiments, we show that the proposed  hierarchical search algorithm is able to find a convolutional cell that achieves a test error of $2.33\%$    on CIFAR-10 and a top-1 error of  $24.17\%$ on IMAGENET for an image classification task. This is the current state-of-the-art result among all DARTS-alike methods based on the original cell-based search space and control of the final training hyper-parameters as in DARTS. We also conducted extensive experiments on four variants of DARTS search spaces and  compare the generalization ability of five  state-of-the-art DARTS-alike algorithm, including DARTS \cite{liu2018darts}, CARS \cite{yang2019cars}, PC-DARTS \cite{xu2019pc}, and DARTS-ES \cite{zela2019understanding} on CIFAR10.  On all these four spaces, our method StacNAS performs the best.

\section{Related Work}
Due to its efficiency and effectiveness for the NAS task, DARTS has attracted a lot of attention. Cai et al. \cite{cai2018proxylessnas} propose ProxylessNAS using path sampling during the search to reduce the high memory cost of DARTS. Realizing the depth gap between the search stage (a shallow network of 8 cells is used) and the final architecture (the discovered cell is repeated to stack a 20 cells structure), Chen et al. \cite{chen2019progressive} proposed PDARTS to progressively prune the search space and deepen the search network. In the analysis we present later, it is shown that this pruning of the search space is harmful. Specifically, it results in an imbalanced number of operators in  different groups: i.e. more Convolution operators than Pooling. PDARTS tries to bridge the depth gap by manually tuning the depth used and controlling the number of skip connections to obtain better performance on CIAR10. In this work, we go one step further. We identify the root cause of the problem as the optimization complexity gap due to use of a shallow proxy structure to search for the cell structure for a deeper final structure. This causes the search stage to converge to a cell with an improper number of skip connections. Arber et al. \cite{zela2019understanding} examine DARTS on several variants of the search spaces, and propose to robustify DARTS by using regularization tricks such as early stopping. In this work, we tackle the robustness problem of DARTS by analyzing two root causes of failure/instability of DARTS, i.e., the linear correlation of similar operators and the optimization complexity gap. 

\section{When and Why DARTS Fails}

\subsection{Preliminaries}
\label{searchspace}
We  use the cell-based search space of DARTS,  where a convolutional neural network is formed by stacking a series of building blocks called cells \cite{zoph2018learning,liu2018darts}. Only two types of cells are  learned: the \textit{normal cell} and  the \textit{reduction cell}. A cell can be seen as a  directed acyclic graph (DAG), where an ordered sequence of $N$ nodes  $\{x_1,\cdots, x_N\}$ are connected by directed edges $(i,j)$.  Each node $x_i$ is a latent representation (i.e. a feature map)  and each directed edge $(i,j)$ represents some operator $o(\cdot) \in O$ that transforms $x_i$.  Within a cell, each internal node is computed as the sum of  all  its predecessors: $
x_j=\sum_{i<j} o^{(i,j)}  (x_i)$.  The task of designing the cell structure is to determine the most important two preceding edges for all internal nodes and the best choice of operators for these selected edges.

DARTS \cite{liu2018darts} proposed a continuous relaxation of the categorical choice of operators and edges so that their relative importance can be learned through stochastic gradient descent (SGD). Specifically, to make the search space continuous, DARTS replaces the discrete choice of operator $o(\cdot) \in O$ with a weighted sum  over all candidate operators $
     \bar{o}^{(i,j)} (x) =\sum_{o \in O}  \alpha_{o}^{(i,j)} o^{(i,j)}  (x_i) $
where $
     \alpha_{o}^{(i,j)}=\frac{\mbox{exp}( \beta_o^{(i,j)})}{\sum_{o' \in O } \mbox{exp}( \beta_{o'}^{(i,j)})} $  is called the architecture parameters. 


After the relaxation,  the task of architecture search reduces to learning a set of continuous architecture parameters $\alpha=\{\alpha^{(i,j)}\}$.    Specifically, one first trains a proxy supernet that contains all candidate operators and connections/edges.  During training, one would be able to learn which operators and edges are redundant. After the training algorithm converges, a compact architecture is obtained by pruning the unimportant operators and edges. 


In DARTS, \cite{anandalingam1992hierarchical} formulate the optimization of the architecture parameters $\alpha$ and the weights $w$ associated with the architecture  as a bi-level optimization problem, where $\alpha$ is treated as the upper-level variable and $w$ as the lower-level variable: 
\begin{equation}
\min \limits_{\alpha} \mbox{ } L_{val} (w^*(\alpha), \alpha), \qquad
  \mbox{   s.t.   } w^*(\alpha) = \mbox{argmin}_w \mbox{  } L_{train} (w,\alpha),
\label{eqn:twolevel1}
\end{equation}
where $L_{train}$ and $L_{val}$ denote the  training and validation loss respectively. 

\subsection{Examining DARTS on Different Search Spaces}
\label{sec:problem}
\paragraph{The Failure Pattern} To examine the generalization problem of the DARTS-alike algorithms, we  use the following five search spaces \textbf{S1} to \textbf{S5}. \textbf{S1} is the original DARTS search space where each type of operators has two members except for \textit{SkipConnect}. For example, there are two pooling operators and two SepConv operators. To examine the correlation problem,  in \textbf{S2}, we add more pooling to the search space and in \textbf{S3} we add more convolution to the search space.  In \textbf{S4},  all poolings are removed and in \textbf{S5}  we use a search space with  the most frequent operators in the searched cells reported by \cite{liu2018darts} to mimic the progressive pruning process proposed in \cite{chen2019progressive}:

\begin{itemize}
\item \textbf{S1:}   \{\textit{MaxPooling $3\times3$, AvePooling $3\times3$, SepConv $3\times3$, SepConv $5\times5$,  DilConv  $3\times3$, DilConv $5\times5$, SkipConnect, Zero}\}

\item \textbf{S2:}   \textit{\{MaxPooling $3\times3$, MaxPooling $5\times5$, AvePooling $3\times3$, AvePooling $5\times5$, SepConv$3\times3$, SepConv $5\times5$,    SkipConnect, Zero}\} 

\item \textbf{S3:}   \{ \textit{MaxPooling $3\times3$,   SepConv$3\times3$, SepConv $5\times5$,   SepConv $7\times7$, DilConv  $3\times3$, DilConv $5\times5$, SkipConnect, Zero} \} 

\item \textbf{S4:}   \{\textit{SepConv$3\times3$, SepConv $5\times5$,   DilConv  $3\times3$, DilConv $5\times5$, SkipConnect, Zero}\} 

\item \textbf{S5:}  \{\textit{ SepConv$3\times3$,  SkipConnect, Zero}\} 
\end{itemize}
Note that  Zero operator here serves as a scaling of each edge, allowing the scoring of the contribution of each edge.    
\begin{table}[htbp]
\centering
\caption {Results of DARTS on various search space base on CIFAR10. Random Baseline is obtained by evaluating 4 randomly generated architectures from the search spaces. }
\begin{tabular}{cccc}
\hline
Search Space & Random Baseline & DARTS Test Error & Characteristics (normal cell) \\
\hline
S1 original          & $96.90\pm0.23$      &  $97.03\pm0.32$     &  Mixed \\ 
S2 more pooling      & $96.59\pm0.60$      &  $96.78\pm0.20$     &  All operators are SepConv \\
S3 more convolution  & $97.14\pm0.16$      &  $91.87\pm3.30$     &  All operators are SkipConnect \\
S4 no pooling        & $96.81\pm0.15$      &  $88.36\pm5.10$     &  All operators are SkipConnect \\
S5 less convolution  & $97.31\pm0.15$      &  $95.67\pm1.20$     &  All operators are SkipConnect \\
\hline
\end{tabular}

 \label{tab:DARTSS1S5}
\end{table}

By examining DARTS on search space S1-S5, we found that DARTS performs poorer than the random baseline on three of them (see Table \ref{tab:DARTSS1S5}), which means, the search process is not only a waste of GPU time but also results in poorer architecture than a randomly generated one. By taking a closer look at the characteristics of the cells found by DARTS, we see that when the search space contains more pooling, DARTS tends to find a cell dominated with Convolution (Sep or Dil); when the search space contains more convolution, DARTS tends to end up with a cell dominated with Pooling or SkipConnection. 

\paragraph{Why It Fails}  
This phenomenon shows that when a certain group of operator (for example convolution operator) has more group members, the search algorithm would   less likely  choose it. We hypothesize  that operators from the same group may serve similar functions/contributions to the final prediction, and hence they would compete for the final importance score assigned to this group. Indeed, the construction of the continuous search space as described in Section \ref{searchspace} is very similar to a multiple regression model: feature maps transformed by all candidate operators are summed together and weighted by coefficients $\alpha$. Thus, when operators are producing highly correlated feature maps,  their weights $\alpha$ may no longer be representative of their real importance. However, in DARTS, the relative importance of operator $o$ is learned through the value of  $\alpha_o$. 

For example, consider the search space  S3. Suppose the system assigns SepConv a score of 0.6 and MaxPool a score of 0.3. However, there are three SepConvs in the system serving a similar function, so they would compete with each other and each gets a score of 0.2. Based on the rule of DARTS, MaxPool has a higher importance score of 0.3, so the system would choose MaxPool for this edge. 

To verify this conjecture,  we calculate the correlation between the flattened feature map of two operators for the first edge of the first cell in the converged supernet of DARTS.  The results  presented in Figure \ref{fig:corrmatrix} shows that  the total operators cluster to four groups (see Figure \ref{fig:cluster}). 

\subsection{The Effect of Supernet Depth and Optimization Complexity Gap}
\paragraph{The Failure Pattern } In the supernet of DARTS, all the candidate operators have to be updated at each gradient step, which incurs large memory cost. A typical solution adopted by a DARTS-alike algorithm \cite{liu2018darts, zela2019understanding, xu2019pc} is to shrink the width and the depth of the proxy supernet during the search stage. For example, in DARTS, during the search stage, a proxy supernet with a 16-channel width and an 8-cell depth is used. After the search converges and redundant operators and edges are pruned, the discovered cell is enlarged and repeated to obtain a 36-channel and 20-cell final structure. We examine the performance of DARTS using different numbers of stacked cells for the search stage, and found that the search results  are very sensitive to this depth. As shown in Table \ref{tab:depth}, when  the search stage architecture in the proxy search space is too shallow, the algorithms would tend to select fewer skip connections and when  the search stage architecture is too deep, the algorithm would select more than the optimal number of skip connections. For example, when we increase the number of cells used for the search stage from 5 to 20, the number of skip connections in the normal cell will increase from 0 to 4. 
\begin{table}[t]
\centering
\caption{Results of DARTS on CIFAR10 by using different supernet  depth during search.}
\begin{tabular}{|c|cccc|}
\hline
Supernet depth (No. of cells)   & 5 & 14 & 17 & 20 \\
\hline
No.of SkipConnect in normal cell & 0 & 2  & 3  & 4 \\
\hline
Test Error & $2.95\pm 0.23$ & $2.75 \pm 0.21$& $ 2.91 \pm 0.27$ & $3.14 \pm 0.23$\\
\hline
\end{tabular}

\label{tab:depth}
\end{table}

\paragraph{Why It Fails} We argue that the unstable result of DARTS when using different cells for the search stage than the final structure is due to  \textbf{the optimization complexity gap  between  the  proxy  search  stage  and  the  final architecture}. As is well studied \cite{he2016identity,sankararaman2019impact}, including skip connections would make the gradient flows easier and optimization of the deep neural network more stable.  In DARTS, to save memory,  a shallow network with 8 cells are used for the search stage and a deep network with 20 cells are used for the final architecture. This depth gap causes a huge optimization complexity gap between them, and the system would not be able to foresee the optimization difficulty when the cells are stacked to a much deeper network. An improper number of skip-connections may be selected as a result.  
\subsection{Hierarchical   Search via operator Clustering}
\label{method}
\subsubsection{Two-stage Search via Operator Clustering}
\label{sec:group}
\begin{wrapfigure}[12]{r}{0.55\textwidth}
\centering
 \includegraphics[width=0.32
 \textheight]{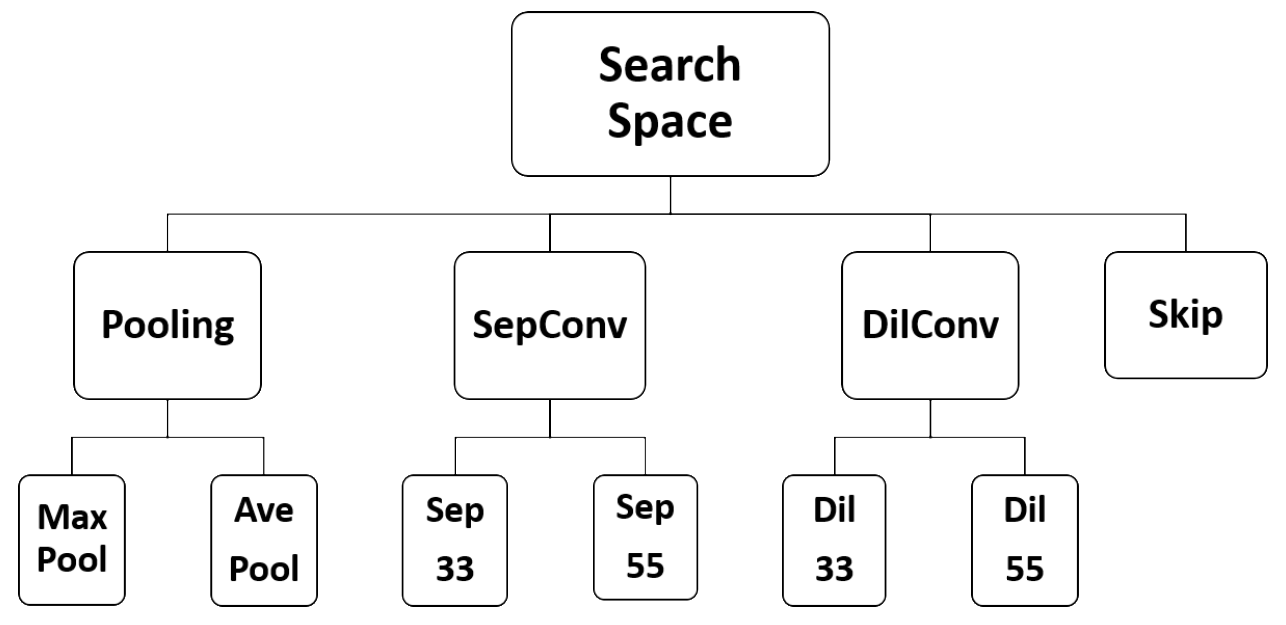}
\caption{Operator clustering diagram: the search space  clusters to four groups.}
\label{fig:cluster}
\end{wrapfigure}

To remedy this problem, we propose a two-stage hierarchical search strategy through   operator clustering. The correlation matrix presented in Figure \ref{fig:corrmatrix} suggests that 
\textit{MaxPool} and \textit{AvePool} are in one group (their correlation is 0.52),  $3 \times 3$  and  $5 \times 5$ \textit{SepConv} are in the same group (correlation 0.38),   $3 \times 3$  and  $5 \times 5$  $DilConv$ are in the same group (correlation 0.27). Therefore, we group all the operators to four clusters as shown in Figure \ref{fig:cluster}. 

\begin{figure}[hbtp]
    \begin{center}
    \includegraphics[width=0.95\linewidth]{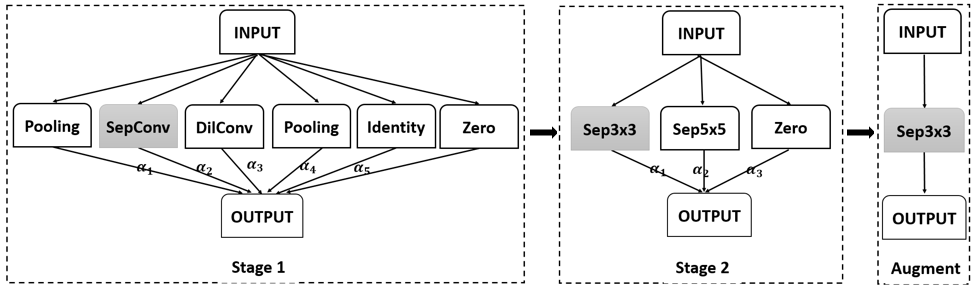}
    \caption{An overview of the algorithm: (a)   Improved DARTS by placing a by-group mixture of candidate operators on each edge. (b) When a certain type of operator is activated, include all candidate operators in this type to form a new mixture operator. (c) Derive the final architecture.  }
    \label{fig:main}
\end{center}
\end{figure}

In the first stage of the algorithm, the best group (for example, SepConv) is activated; and in the second stage, the best operator in the activated group is selected (for example SepConv $3 \times 3$) as shown in Figure  \ref{fig:main}. 
 


\subsubsection{Optimization Complexity Matching}
\label{sec:discrepancy}

As studied in \cite{sankararaman2019impact}, the key factors that affect the optimization difficulty include the width, the depth and the number of SkipConnect. Narrowing the network, deepening the network, and including fewer SkipConnect increase the optimization difficulty. In DARTS, both the width and the depth are changed for the proxy search supernet. Therefore, to select  a proper number of SkipConnect, one must shrink the network in a balanced way. 

To quantify the optimization difficulty more concretely,  we use a criterion called gradient confusion, introduced in \cite{sankararaman2019impact}.  The authors empirically show that when gradient confusion is low, SGD has better convergence properties. Deeper networks usually correspond to higher gradient confusion, and skip connection can significantly reduce gradient confusion.  

Formally, let $\{l_i\}_{i=1}^m$ be losses for mini-batches, and let $w$ be trainable parameters.  \textit{Gradient confusion} is defined to be the maximum value of the negative inner product between gradients evaluated at all mini-batches which is defined by the following formula:
\begin{equation*}
    \zeta =  \max_{i,j = 1,\dots,m}\{-<\nabla l_i(w), \nabla l_j(w)>\}.
    \label{con: our gradient confusion}
\end{equation*}
In Table \ref{tab:confusion}, we present the normalized estimated gradient confusion ($\zeta$ is divided by the number of parameters)  for both search stage and final architecture with different depth and a fixed width of 16 channels (estimation details are provided in the Appendix \ref{sec:confusion}).  We can see that the desired depth for search stage 1 is 14 cells, to make the optimization complexity compatible with that of the final architecture. This matching enables the proposed algorithm   to optimize the search architecture with the same level of difficulty as with the final architecture.  

Empirically, we found that the search results are sensitive to the depth/width design of search stage 1 but are insensitive to the depth/width design of search stage 2. This seems to be due to the fact the number of skip connections are already determined in search stage 1. To facilitate easy use of the algorithm for different memory constraints,  we provide several settings with different depth/width in  Table \ref{tab:confusion} of Section \ref{sec: exp}.
 
\subsubsection{One-level Optimization}

Recent works \cite{liu2020autofis,stamoulis2019single} have shown that one-level optimization performs equally good or even better than the original two-level optimization proposed by DARTS \cite{liu2018darts}. In our ablation study in Section~\ref{sec:abl}, we also verified that coupled with the proposed solution, one-level optimization is superior to two-level optimization with respect to both accuracy and stability.  Therefore, in this work, we adopt the simpler one-level optimization approach, where the parameters are updated simultaneously using: $ w_t=w_{t-1}-\eta_t  \partial_w L_{train}(w_{t-1},\alpha_{t-1})$ and $
  \alpha_t=\alpha_{t-1}-\delta_t  \partial_{\alpha} L_{train}(w_{t-1},\alpha_{t-1})$, where $\eta_t$ and $\delta_t$ are learning rates. 

 
 


\section{Experiments}
\label{sec: exp}
In this section, we validate the effectiveness of our proposed method for the image classification task. 
\subsection{NAS Benchmarks on CIFAR10/CIFAR100}

\paragraph{Implementation Details}  We conduct our first set of experiments on CIFAR-10 and CIFAR-100 \cite{krizhevsky2009cifar}, each containing 50,000 training   images and 10,000 test images. CIFAR-10 has 10 classes and CIFAR-100  has 100. For search space \textbf{S1-S5} (see Section \ref{sec:problem}),   we first estimate the correlation matrix of all the candidate operators to identify the underlying groups for the given data (the estimation method and clustering results of \textbf{S1-S5} is provided in the Appendix Section \ref{Appsec:corr}). Then we select the representative operators\textemdash for fair competition among groups, we prefer to select those with the same kernel size, i.e. $3\times3$\textemdash for each group to obtain the search space of search stage 1. In our ablation study in Section \ref{sec:abl}, we show that the final results are not sensitive to the selection of the representative operators.    

In the first stage,   the proxy parent network  of  14 cells is trained using one-level optimization for 80 epochs.  After convergence, the optimal operator group is activated based on the learned $\alpha$. In the second stage, we replace the mixed operator  with a weighted sum of all operators in the activated group from stage 1. Then a proxy parent network by stacking 20 cells is trained using one-level optimization for 80 epochs. 

When the best cells are identified by the search stage, they are stacked to build the final architecture. We evaluate the final architecture using exactly the same   training scheme like DARTS for a fair comparison. More details   are provided in the Appendix \ref{sec:train_tricks}. 

\paragraph{Search Results} The search results on the original DARTS search  space (\textbf{S1}) are provided in Table \ref{tab:c10bench} for CIFAR10 and Table \ref{tab:c100bench} for CIFAR100. 

We also conducted extensive experiments on   search spaces \textbf{S2-S5} and  compare StacNAS with   five other  well-established DARTS-alike algorithm, including DARTS \cite{liu2018darts}, CARS \cite{yang2019cars}, PC-DARTS \cite{xu2019pc}, and Robust\_DARTS \cite{zela2019understanding} on CIFAR10. As we can see from the results plotted in Figure \ref{fourspaces} (Table is provided in APPENDIX \ref{benchtable}),  our algorithm performs the best across all the search spaces. 
. 
\begin{figure}[htbp]
\caption{Comparison of  five search methods and random sampled architecture (4 seeds) from their respective search spaces. Methods  above the diagonal outperform the random sampled ones, vice versa. Left: CIFAR10 benchmark   across search space S2-S5; Right: Some results of DARTS excluded for better demonstration.  }
\begin{tabular}{cc}
\begin{minipage}[c]{0.5\linewidth}
  \centering
    \includegraphics[width=6cm]{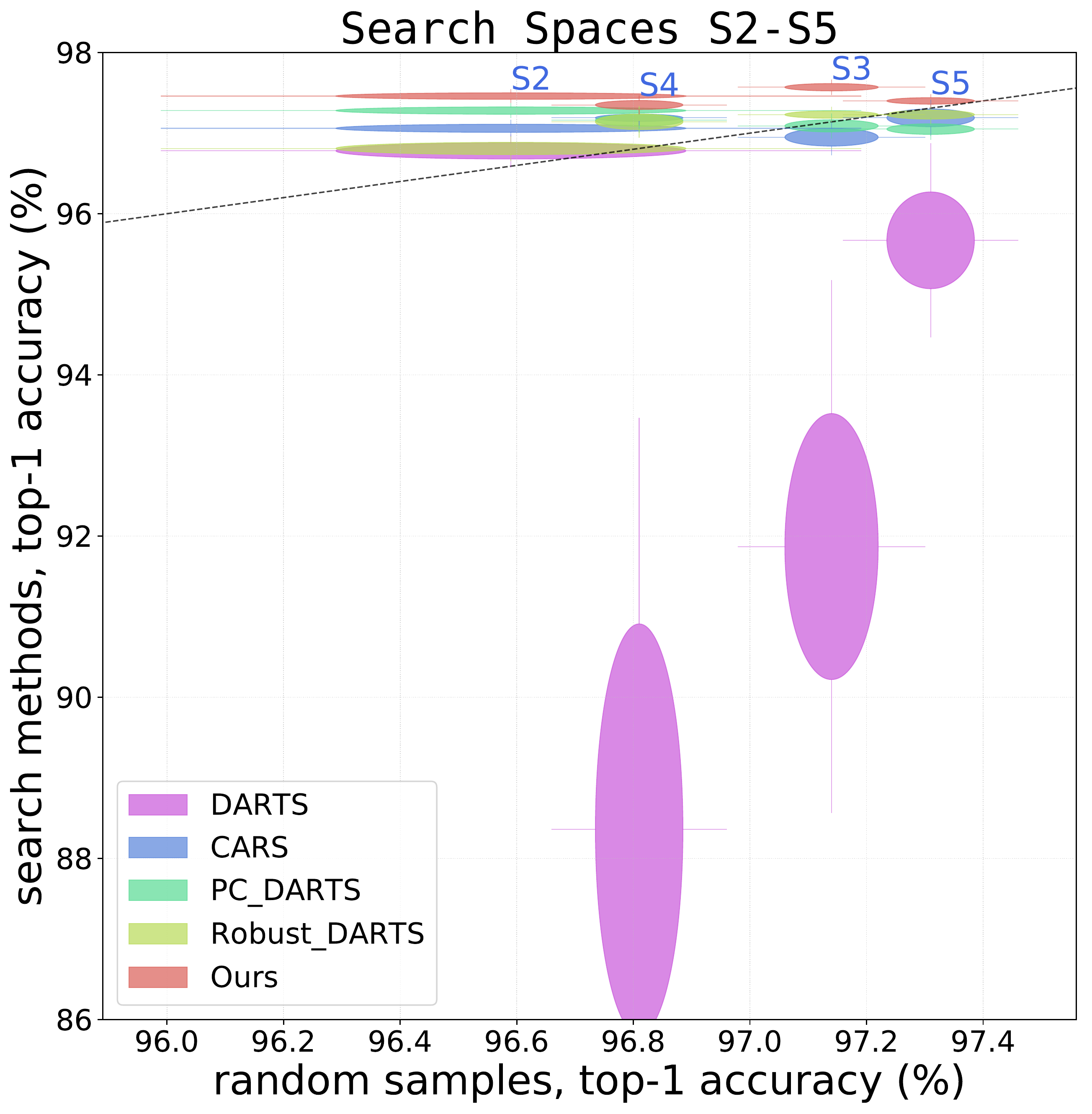} 
     
\end{minipage} 
\begin{minipage}[c]{0.5\linewidth}
  \centering
    \includegraphics[width=6 cm]{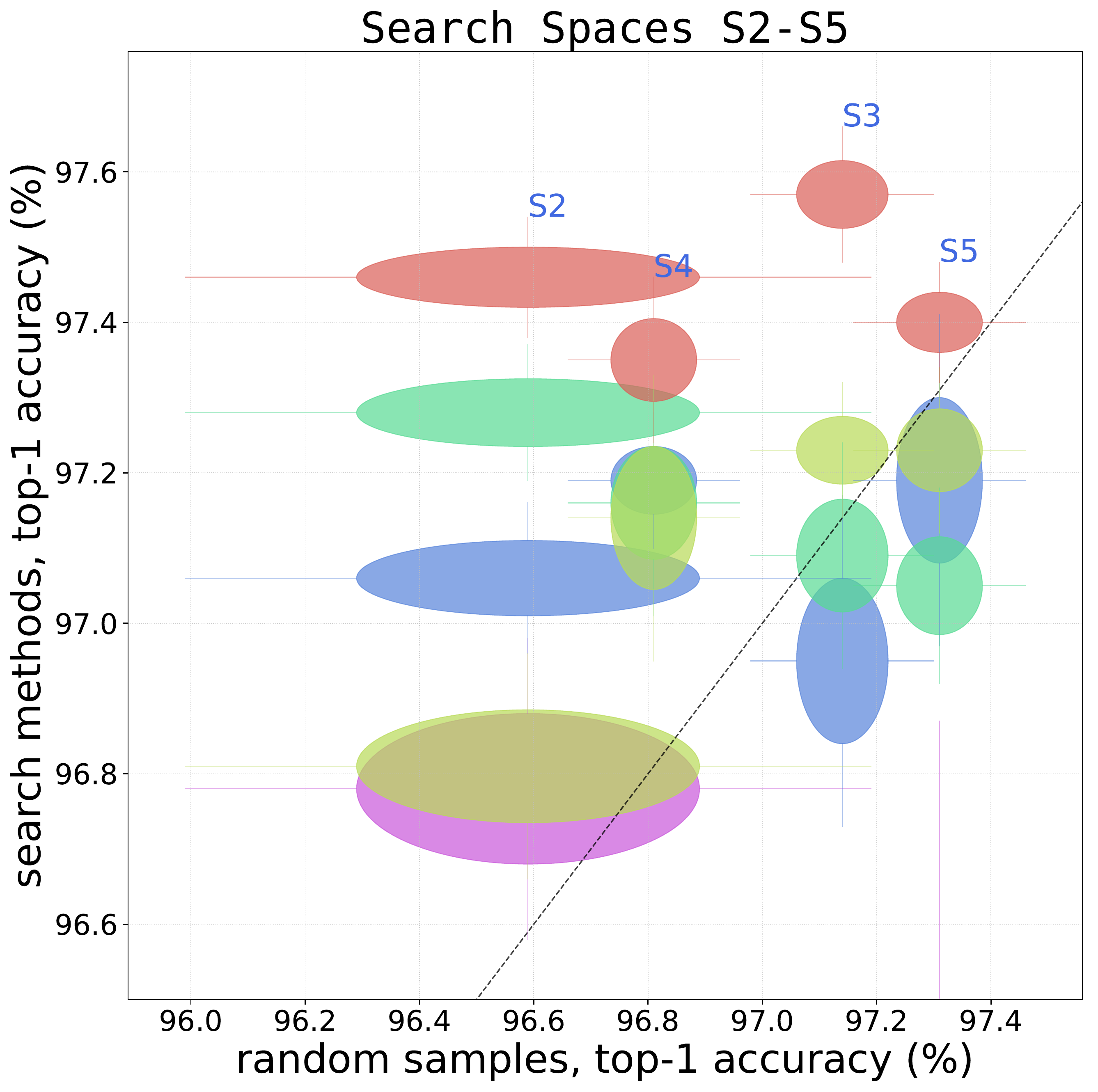} 
\end{minipage}
\end{tabular}

\label{fourspaces}
\end{figure}

\begin{table}[htbp]
\begin{tabular}{cc}
\begin{minipage}[c]{0.52\linewidth}
\caption {CIFAR-10 Benchmark based on cell-based search space. * Proxyless reproduced by us on the cell-based search space.}
\centering
\begin{tabular}{ccc}
\hline
 Method   &  Test Error (\%)  & \small \small Optim  \\
 \hline
 ENAS\cite{pham2018efficient} & 2.89 & RL  \\
 CARS \cite{yang2019cars} & 2.62 & Evolution   \\ 
 DARTS \cite{liu2018darts}    & 2.76$\pm$0.09 & Gradient     \\
 \small Proxyless\cite{cai2018proxylessnas}*& $2.8\pm 0.2$ & Gradient\\
 PDARTS \cite{chen2019progressive} & 2.5 & Gradient \\
\small\small PC-DARTS  \cite{xu2019pc}   & 2.57 $\pm$0.07 & Gradient    \\
\small \small Robust\_DARTS \cite{zela2019understanding} & $2.95\pm0.21$  & Gradient  \\
\hline
\small StacNAS      & 2.48$\pm$0.08   &  Gradient          \\
\hline
\end{tabular}

 \label{tab:c10bench}
\end{minipage}
\begin{minipage}[c]{0.45\linewidth}
\caption{CIFAR-100 Benchmark: $^\star$ performed by us. }
\centering
\begin{tabular}{cc}
\hline
Method    &  Test Error (\%)      \\ \hline

\small DARTS$^\star$(base)   \cite{liu2018darts}          & 17.63 $\pm$ 0.67             \\
\small PDARTS (base)  \cite{chen2019progressive} & 15.92     \\
\small \small Robust\_DARTS \cite{zela2019understanding} & 18.01 $\pm$ 0.26 \\
\hline
\small StacNAS      & 15.41$\pm$0.2   \\
\hline
\end{tabular}
\centering
 \setcaptionwidth{2in} 

\label{tab:c100bench}
\end{minipage}
\end{tabular}
\end{table}
\begin{table*}[htb]
\centering
\caption{ImageNet Benchmark (mobile setting).    IMAGENET(base) represents that the architecture are   trained the same  as DARTS. (AA) represents that the architecture  are trained with AutoAugmentation.}
\begin{tabular}{cccccccc}
\hline
                              & \multicolumn{2}{c}{Test Error (\%)}      & Params & GPU  & Search   & Search Space      \\
Architecture           & top-1           & top-5 & (M)    & Days  & Method         \\
\hline
AmoebaNet-C \cite{real2019regularized}      & 24.3           & 7.6   & \textbf{6.4}    & 3150        & evolution &   Cell-based    \\
DARTS   \cite{liu2018darts}      & 26.7            & 8.7   & 4.7    & N/A          & SGD & Cell-based \\
PDARTS  \cite{chen2019progressive} & 24.4 & 7.4 & 4.9 & N/A  &SGD & Cell-based\\
PC-DARTS \cite{xu2019pc} & 24.2 & 7.3 & 5.3 & 3.8 &SGD & Cell-based\\
EfficientNet-B0  \cite{tan2019efficientnet} & 23.7 & 6.8 & 5.3 & N/A & RL & \small MobileNet Backbone\\ 

\hline

StacNAS (base)                  &      24.17      & 6.9  & 5.7   &  20         & SGD & Cell-based \\
StacNAS   (AA)                &     \textbf{23.12}         & 5.9 & 5.7   & 20         & SGD & Cell-based\\

                    \hline
\end{tabular}
\label{tab:imagenet}
\end{table*}

\subsection{NAS Benchmarks on IMAGENET}
We directly search using the whole IMAGENET dataset without subsampling on the original DARTS search space. In the first stage, 11 cells and 22 channels are used for the proxy search supernet; in the second stage, 14 cells and 22 channels are used. After the search converges, the discovered cells are stacked to form a 14 cells and 48 channel network.  As shown in  Table \ref{tab:imagenet}, we achieve a top-1  error of 23.12\%, which is the best-known performance to date based on the cell-based search space. Other training details are provided in Appendix \ref{sec:train_tricks}. 
\subsection{Ablation Experiments}
\label{sec:abl}
\textbf{Effectiveness of operator clustering}: As shown in Figure \ref{fourspaces}, although some other  DARTS-alike algorithms fail to outperform the random baseline under some unbalanced search spaces, our operator clustering-based hierarchical search algorithm performs always beat the random baseline as well as the other algorithms across all search spaces. Especially, most of the DARTS-alike algorithms would discover cells dominated by \textit{SkipConnect} or \textit{Sep3$\times$3} (Appendix \ref{sec:cell}),  while our algorithm avoided this failure and found well-mixed cells  (Appendix \ref{sec:cell}). 

\textbf{Effectiveness of complexity match via gradient confusion}: in this part, we examine the effectiveness of  complexity match by comparing the performance of the proposed algorithm under different width/depth combination settings on CIFAR10: 
\begin{table}[htbp] 
\small
\caption{Complexity match: different width (channels)/depth (cells) settings for search stage 1.}
\begin{tabular}{|c|c|ccc|ccc|}
\hline
                 & Final & Match1 & Match2 & Match3 & Poor1 & Poor2 & Poor3 \\
                 \hline
Channels & 36    & 32              & 16              & 10 & 16  & 16    & 16         \\
Cells   & 20    & 17              & 14              & 11 & 5  & 17  & 20 \\
Confusion & \textbf{0.57} & \textbf{0.58} & \textbf{0.56} & \textbf{0.55} & 0.36 & 0.87 & 1.02 \\
Error (\%) & -  & \small 2.52 $\pm$ 0.09 &  \small 2.48 $\pm 0.08$ & \small 2.5 $\pm 0.08$ & \small 2.87 $\pm$ 0.09 & \small 2.95 $\pm$ 0.12 &  \small 3.12 $\pm$ 0.15 \\
\small No\_Skip & - & 2 & 2 & 2 & 0 &  3 & 4 \\ 
\hline
\end{tabular}

\label{tab:confusion}
\end{table}

\textbf{Contribution of each component}: In Table \ref{ab}, we conduct experiments to check how much each

of the proposed components including hierarchical  search via clustering, complexity matching, and  one-level optimization contributes to the final performance. 

\begin{table} [htbp] 
\centering
  \caption{Ablation Experiments. * This  result is reproduced by us using the publically available code.}
\begin{tabular}{ccc}
\hline
                              & \multicolumn{2}{c}{CIFAR10}                                \\
                              & two-level & one-level       \\
                              \hline
DARTS baseline              & 2.97$\pm$0.32*             & 2.74$\pm$0.12    \\
operator clustering  & 2.82$\pm$0.26             & 2.68$\pm$0.10    \\
\small clustering+depth match & 2.73$\pm$0.23             & 2.48$\pm$0.08 \\
\hline
\end{tabular}
\label{ab}
\end{table}



\textbf{Selection of representative operators}:  We also experiment on CIFAR10 with randomly selected operators from each group for stage 1. The results for CIFAR10 are $2.53\pm 0.08$, which means the selection of the operators for the first stage  is not sensitive.

\textbf{Predicted Performance Correlation (CIFAR-10)}
In DARTS, the learned architecture parameter $\alpha$  is supposed to represent the relative importance of one candidate operator verse the others. In this part, 2e check the correlation between the   accuracy of a stand-alone architecture with different candidate operators and the corresponding $\alpha$. For the original DARTS algorithm, this correlation is as low as only 0.2, while our algorithm, achieves a  surprisingly high correlation of 0.9. This results provide strong evidence what the properly learned $\alpha$ in the differential NAS framework is trustworthy. Detailed implementation and correlation results are provided in Appendix \ref{ab:stand_alone}.

\section{Conclusion}
This paper introduced a  hierarchical  search solution via operator clustering  to differentiable NAS. Our method addresses some difficulties encountered in the original DARTS algorithm such as  grouping of correlated operators, and the fundamental challenges of matching neural network optimization complexity in NAS and in the final training. It was shown that our method leads to the state-of-the-art image classification results on multiple benchmark datasets and search spaces.

\bibliography{references.bib}

\begin{thebibliography}{10}

\bibitem{zoph2016neural}
Barret Zoph and Quoc~V Le.
\newblock Neural architecture search with reinforcement learning.
\newblock {\em ICLR}, 2017.

\bibitem{bender2018understanding}
Gabriel Bender, Pieter-Jan Kindermans, Barret Zoph, Vijay Vasudevan, and Quoc
  Le.
\newblock Understanding and simplifying one-shot architecture search.
\newblock In {\em ICML}, pages 549--558, 2018.

\bibitem{liu2018darts}
Hanxiao Liu, Karen Simonyan, and Yiming Yang.
\newblock {DARTS}: Differentiable architecture search.
\newblock In {\em ICLR}, 2019.

\bibitem{guo2019single}
Zichao Guo, Xiangyu Zhang, Haoyuan Mu, Wen Heng, Zechun Liu, Yichen Wei, and
  Jian Sun.
\newblock Single path one-shot neural architecture search with uniform
  sampling.
\newblock {\em arXiv preprint arXiv:1904.00420}, 2019.

\bibitem{xie2018snas}
Sirui Xie, Hehui Zheng, Chunxiao Liu, and Liang Lin.
\newblock {SNAS}: stochastic neural architecture search.
\newblock In {\em International Conference on Learning Representations}, 2019.

\bibitem{cai2018proxylessnas}
Han Cai, Ligeng Zhu, and Song Han.
\newblock Proxyless{NAS}: Direct neural architecture search on target task and
  hardware.
\newblock In {\em ICLR}, 2019.

\bibitem{yang2019cars}
Zhaohui Yang, Yunhe Wang, Xinghao Chen, Boxin Shi, Chao Xu, Chunjing Xu,
  Qi~Tian, and Chang Xu.
\newblock Cars: Continuous evolution for efficient neural architecture search.
\newblock {\em CVPR}, 2020.

\bibitem{zela2019understanding}
Arber Zela, Thomas Elsken, Tonmoy Saikia, Yassine Marrakchi, Thomas Brox, and
  Frank Hutter.
\newblock Understanding and robustifying differentiable architecture search.
\newblock In {\em ICLR}, 2020.

\bibitem{sankararaman2019impact}
Karthik~A Sankararaman, Soham De, Zheng Xu, W~Ronny Huang, and Tom Goldstein.
\newblock The impact of neural network overparameterization on gradient
  confusion and stochastic gradient descent.
\newblock {\em arXiv preprint arXiv:1904.06963}, 2019.

\bibitem{xu2019pc}
Yuhui Xu, Lingxi Xie, Xiaopeng Zhang, Xin Chen, Guo-Jun Qi, Qi~Tian, and
  Hongkai Xiong.
\newblock Pc-darts: Partial channel connections for memory-efficient
  architecture search.
\newblock In {\em International Conference on Learning Representations}, 2020.

\bibitem{chen2019progressive}
Xin Chen, Lingxi Xie, Jun Wu, and Qi~Tian.
\newblock Progressive differentiable architecture search: Bridging the depth
  gap between search and evaluation.
\newblock In {\em ICCV}, pages 1294--1303, 2019.

\bibitem{zoph2018learning}
Barret Zoph, Vijay Vasudevan, Jonathon Shlens, and Quoc~V Le.
\newblock Learning transferable architectures for scalable image recognition.
\newblock In {\em Proceedings of the IEEE conference on computer vision and
  pattern recognition}, pages 8697--8710, 2018.

\bibitem{anandalingam1992hierarchical}
G~Anandalingam and Terry~L Friesz.
\newblock Hierarchical optimization: An introduction.
\newblock {\em Annals of Operations Research}, 34(1):1--11, 1992.

\bibitem{he2016identity}
Kaiming He, Xiangyu Zhang, Shaoqing Ren, and Jian Sun.
\newblock Identity mappings in deep residual networks.
\newblock In {\em European conference on computer vision}, pages 630--645.
  Springer, 2016.

\bibitem{liu2020autofis}
Bin Liu, Chenxu Zhu, Guilin Li, Weinan Zhang, Jincai Lai, Ruiming Tang,
  Xiuqiang He, Zhenguo Li, and Yong Yu.
\newblock Autofis: Automatic feature interaction selection in factorization
  models for click-through rate prediction.
\newblock {\em arXiv preprint arXiv:2003.11235}, 2020.

\bibitem{stamoulis2019single}
Dimitrios Stamoulis, Ruizhou Ding, Di~Wang, Dimitrios Lymberopoulos, Bodhi
  Priyantha, Jie Liu, and Diana Marculescu.
\newblock Single-path nas: Designing hardware-efficient convnets in less than 4
  hours.
\newblock {\em arXiv preprint arXiv:1904.02877}, 2019.

\bibitem{krizhevsky2009cifar}
Alex Krizhevsky, Vinod Nair, and Geoffrey Hinton.
\newblock Cifar-10 and cifar-100 datasets.
\newblock {\em URl: https://www. cs. toronto. edu/kriz/cifar. html}, 6, 2009.

\bibitem{pham2018efficient}
Hieu Pham, Melody Guan, Barret Zoph, Quoc Le, and Jeff Dean.
\newblock Efficient neural architecture search via parameters sharing.
\newblock In {\em ICML}, pages 4095--4104, 2018.

\bibitem{real2019regularized}
Esteban Real, Alok Aggarwal, Yanping Huang, and Quoc~V Le.
\newblock Regularized evolution for image classifier architecture search.
\newblock In {\em Proceedings of the aaai conference on artificial
  intelligence}, volume~33, pages 4780--4789, 2019.

\bibitem{tan2019efficientnet}
Mingxing Tan and Quoc~V Le.
\newblock Efficientnet: Rethinking model scaling for convolutional neural
  networks.
\newblock {\em arXiv preprint arXiv:1905.11946}, 2019.

\bibitem{cubuk2018autoaugment}
Ekin~D Cubuk, Barret Zoph, Dandelion Mane, Vijay Vasudevan, and Quoc~V Le.
\newblock Autoaugment: Learning augmentation policies from data.
\newblock {\em arXiv preprint arXiv:1805.09501}, 2018.

\bibitem{hu2018squeeze}
Jie Hu, Li~Shen, and Gang Sun.
\newblock Squeeze-and-excitation networks.
\newblock In {\em Proceedings of the IEEE conference on computer vision and
  pattern recognition}, pages 7132--7141, 2018.

\end{thebibliography}
\bibliographystyle{unsrt}

\newpage

\appendix
\section{Operator Correlation Estimation}
\label{Appsec:corr}

We Compute the correlations among different operators using CIFAR-10. This is for the purpose of grouping similar operators in our procedure. The correlation measures are obtained as follows.
\begin{enumerate}
    \item A network of 14 cells is trained 80 epochs using parameters in the original DARTS, where the cell DAG consists four intermediate nodes and fourteen learnable edges, and each edge is a weighted summation of all the candidate operators in the space.
    \item With the trained model, for a given training image, we may pass it through the network, and store the output of the candidate operators (exclude "none" and "skip\_connect") on all edges.
     On each edge, flatten the  output tensors into  vectors.
    Repeat until more than 10000 images in the training set have been processed. 
    \item Use the resulting data to calculate the correlations among the operators.
\end{enumerate}
The clustering results for S1 to S5 are listed in the following table:
\begin{table}[htbp]
\centering
 \begin{tabular}{|c|c|c|}
\hline
S1    & Key Operator & \small \small Other Members \\
\hline
Group1 & MaxPool $3\times3$ &  AvePool $3\times3$ \\
Group2 & SkipConnect & -\\
Group3 & SepConv $3\times3$ &  SepConv $5\times5$  \\
Group4 & DilConv $3\times3$ &  DilConv $5\times5$\\
\hline
\end{tabular}
\end{table}
 \begin{table}[htbp]
 \centering
 \begin{tabular}{|c|c|c|}
\hline
S5     & Key Operator & \small \small Other Members \\
\hline
Group1 & SkipConnect & -\\
Group2 & SepConv $3\times3$ & -\\
\hline
\end{tabular}
\end{table}

\begin{table}[htbp]
\centering
\begin{tabular}{|c|c|c|}
\hline
S3     & Key Operator & Other Group Members \\
\hline
Group1 & MaxPool $3\times3$ &  - \\
Group2 & SkipConnect & -\\
Group3 & SepConv $3\times3$ &  SepConv $5\times5$,  SepConv $7\times7$  \\
Group4 & DilConv $3\times3$ &  DilConv $5\times5$\\
\hline
\end{tabular}
\end{table}
\begin{table}[ht]
\centering
\begin{tabular}{|c|c|c|}
\hline
S4     & Key Operator & Other Group Members \\
\hline
Group1 & SkipConnect & -\\
Group2 & SepConv $3\times3$ & SepConv $5\times5$, DilConv $3\times3$, DilConv $5\times5$\\
\hline
\end{tabular}
\end{table}
\begin{table}[h!t]
\begin{tabular}{|c|c|c|}
\hline
S2     & Key Operator & Other Group Members   \\
\hline
Group1 & SkipConnect  &  MaxPool $3\times3$, MaxPool $5\times5$, AvePool $3\times3$,AvePool $5\times5$ \\
Group2 & SepConv $3\times3$ &  SepConv $5\times5$   \\
\hline
\end{tabular}
\end{table}
 \section{Train Details}
 \label{sec:train_tricks}
\textbf{CIFAR-10 and CIFAR-100} The final architecture  is a stack of    20 cells: 18 normal cells and 2 reduction
cells, posioned at the 1/3 and 2/3 of the  network  respectively.   

\textbf{ImageNet}    The final architecture  is a stack of    14 cells: 12 normal cells and 2 reduction
cells, posioned at the 1/3 and 2/3 of the  network  respectively.  For the \textit{base} training,  we trained the network exactly the same with DARTS. For the \textit{AA} training, we add AutoAugment (IMAGENET) \cite{cubuk2018autoaugment} policy and SE \cite{hu2018squeeze}, then trained the network for 600 epochs, other than these, all the other settings are same as DARTS. 
\section{Esimation of Gradient Confusion}
 \label{sec:confusion}
We compared gradient confusion for several different settings, including different numbers of cells, different stages, with/without an auxiliary head. The result are provided in the following table. 

For each setting, we first train the architecture for 100 epochs and then ran 10 more iterations with randomly selected mini-batch using the trained parameters and compute the gradient confusion using formula (\ref{con: our gradient confusion}) and obtain $ \zeta_i$ for each iteration $i$. Then, we compute the mean of the 10 $\zeta_i$ we got and divided them by the corresponding model parameter size $M$. Formally, \begin{equation}
    \Bar{ \zeta}=\sum_{i=11}^{10}  \zeta_i/M
\end{equation}
We treat it as the  measurement of the optimization complexity  for the corresponding setting.  
\section{Predicted Performance Correlation}
\label{ab:stand_alone}
\begin{figure}[htbp]
\begin{tabular}{cc}
\begin{minipage}[c]{0.5\linewidth}
  \includegraphics[width=7cm]{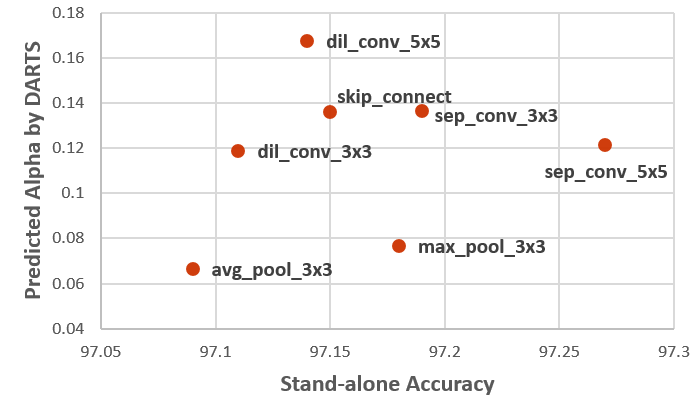} 
       
\end{minipage}
\begin{minipage}[c]{0.5\linewidth}
\centering
  \includegraphics[width=7cm]{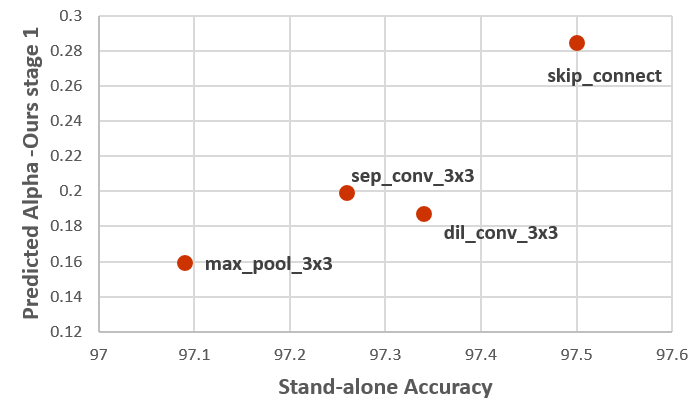} 
        
\end{minipage}
\end{tabular}
\caption{Correlation between standalone model and learned $\alpha$: (a) Correlation coefficient of  StacNAS: 0.91; (b) Correlation coefficient of DARTS: 0.2. }
\label{fig:corr}
\end{figure}

In DARTS, the learned architecture parameter $\alpha$  is supposed to represent the relative importance of one candidate operator verse the others. To check the correlation between the   accuracy of a stand-alone architecture with different candidate operators and the corresponding $\alpha$, we replace the selected operator in the first edge of the first cell for the final architecture with all the other candidate operators,  and fully train them until converge. The obtained stand-alone accuracy is compared with the corresponding $\alpha$ learned for each candidate operator. Their correlation is plotted in Figure \ref{fig:corr}.  

 \section{The Algorithm}
   \begin{algorithm}[htbp]
    \caption{Hierarchical NAS via operator Clustering}\label{euclid}
    \begin{algorithmic}[1]
        \Procedure{Stage 1 }{Group search}
        \State  Create a by-group  operator $\tilde{o}^{(i,j)} (x)$ parametrized by  $\tilde{\alpha}_{o}^{(i,j)}$ for each edge $(i,j)$
        \If { $t <  T $} 
        \State  Update weights $w_t$ by descending $\partial_w L_{train}(w_{t-1},\tilde{\alpha}_{t-1})$
        \State  Update architecture parameter $\tilde{\alpha_t}$ by descending $\partial_{\tilde{\alpha}} L_{train}(w_{t-1},\tilde{\alpha}_{t-1})$
        \EndIf
        \State Set optimal $\tilde{\alpha}^*=\alpha^T$
        \State Activate the optimal operator group  based on    $\tilde{\alpha}^*$
        \EndProcedure
        
    \Procedure{Stage 2 }{Backward}
       \State  Use all the group member of  $o^{*(i,j)}$  to create mixed operator $\hat{o}^{(i,j)} (x)$   parametrized by  $\hat{\alpha}_{o}^{(i,j)}$ 
        \If { $t <  T $} 
        \State Update weights $w_t$ by descending $\partial_w L_{train}(w_{t-1},\hat{\alpha}_{t-1})$
        \State Update architecture parameter $\alpha_t$ by descending $\partial_{\hat{\alpha}} L_{train}(w_{t-1},\hat{\alpha}_{t-1})$
\EndIf
\State Set optimal $\hat{\alpha}^*=\alpha_T$
 \State Select the best operator $o^{*(i,j)} \gets \mbox{argmax}_{o\in O} \hat{\alpha}^{(i,j)}$ 
 
       \EndProcedure

\Procedure{Stage 3 }{Final Architecture}
  \State     Prune the redundant operators and edges.
  
  \EndProcedure          
    \end{algorithmic}
    \label{alg:StacNAS}
\end{algorithm}

\section{CIFAR10 Benchmark Table}
\label{benchtable}
\begin{table}[htbp]
\begin{tabular}{c|cccccc}
\hline
\textbf{} & \textbf{Random} & \textbf{DARTS} & \textbf{CARS} & \small \textbf{PC\_DARTS} & \small \small \textbf{Robust\_DARTS} & \textbf{StacNAS} \\
\hline
\textbf{S2} & 96.59 $\pm$ 0.60 & 96.78 $\pm$ 0.20 & 97.06 $\pm$ 0.10 & 97.28 $\pm$ 0.09 & 96.81 $\pm$ 0.15 & 97.46 $\pm$ 0.08 \\
\textbf{S3} & 97.14 $\pm$ 0.16 & 91.87 $\pm$ 3.30 & 96.95 $\pm$ 0.22 & 97.09 $\pm$ 0.15 & 97.23 $\pm$ 0.09 & 97.57 $\pm$ 0.09 \\
\textbf{S4} & 96.81 $\pm$ 0.15 & 88.36 $\pm$ 5.10 & 97.19 $\pm$ 0.09 & 97.16 $\pm$ 0.15 & 97.14 $\pm$ 0.19 & 97.35 $\pm$ 0.11 \\
\textbf{S5} & 97.31 $\pm$ 0.15 & 95.67 $\pm$ 1.20 & 97.19 $\pm$ 0.22 & 97.05 $\pm$ 0.13 & 97.23 $\pm$ 0.11 & 97.40 $\pm$ 0.08 \\
\hline
\end{tabular}
\caption {Results of different search algorithms on various search space base on CIFAR10. Means and stds are obtained by repeated experiments with 4 seeds. }
 \label{tab:ALLS1S5}
\end{table}

\section{Cells of Benchmark Dataset}
\label{sec:cell}

\begin{figure}[htbp]
  \centering
 \includegraphics[width=8cm]{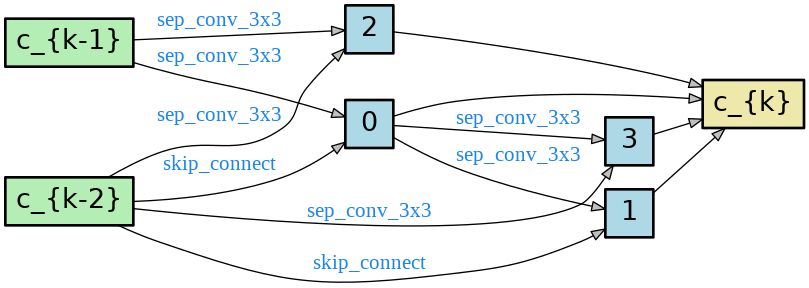} 
       \caption{Normal cell   learned on CIFAR-10 for S1 (StacNAS)}%
 \end{figure}
\begin{figure}[htbp]
  \centering
   \includegraphics[width=12cm]{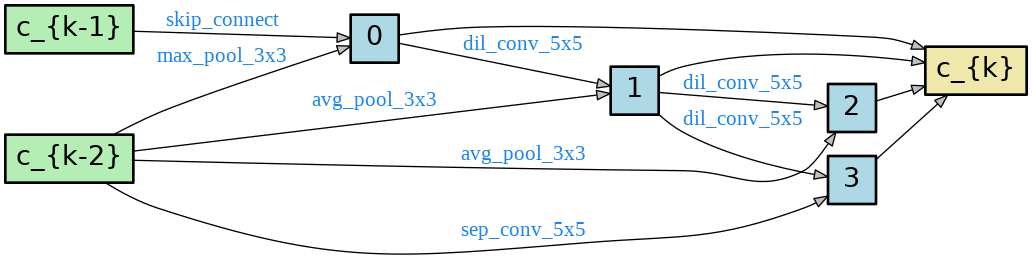} 
       \caption{Reduction cell of CIFAR-10 for S1 (StacNAS)}%
\end{figure}

\begin{figure}[htbp]
\begin{tabular}{cc}
\begin{minipage}[c]{0.5\linewidth}
  \centering
    \includegraphics[width=6cm]{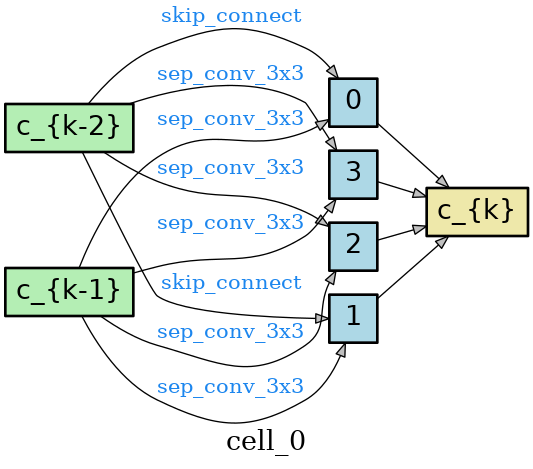} 
       \caption{Normal cell of CIFAR-10 for S2 (StacNAS) }%
     
\end{minipage} 
\begin{minipage}[c]{0.5\linewidth}
  \centering
   \includegraphics[width=6cm]{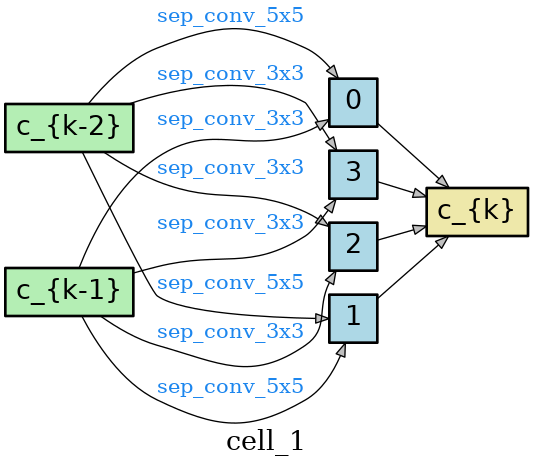} 
       \caption{Reduction cell of CIFAR-10 for S2 (StacNAS) }%
\end{minipage}
\end{tabular}
 
\end{figure}
\begin{figure}[htbp]
\begin{tabular}{cc}
\begin{minipage}[c]{0.5\linewidth}
  \centering
    \includegraphics[width=4cm]{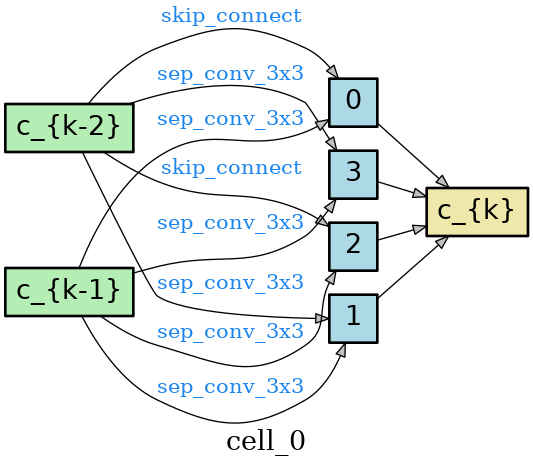} 
       \caption{Normal cell of CIFAR-10 for S3 (StacNAS) }%
     
\end{minipage} 
\begin{minipage}[c]{0.5\linewidth}
  \centering
   \includegraphics[width=6cm]{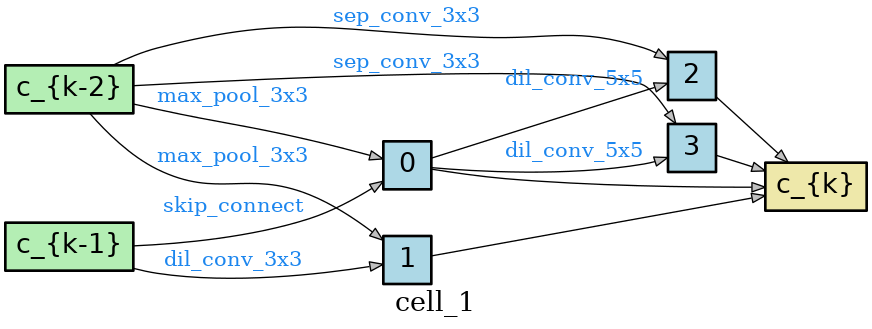} 
       \caption{Reduction cell of CIFAR-10 for S3 (StacNAS) }%
\end{minipage}
\end{tabular}
  
\end{figure}
\begin{figure}[htbp]
  \centering
    \includegraphics[width=8cm]{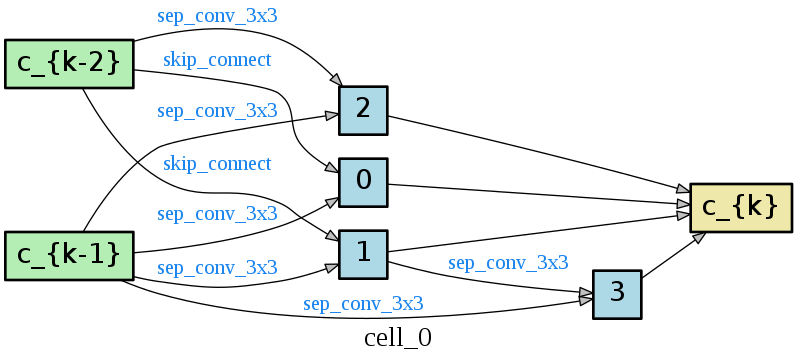} 
       \caption{Normal cell of CIFAR-10 for S4 (StacNAS) }%
\end{figure}

\begin{figure}[htbp]
  \centering
   \includegraphics[width=10cm]{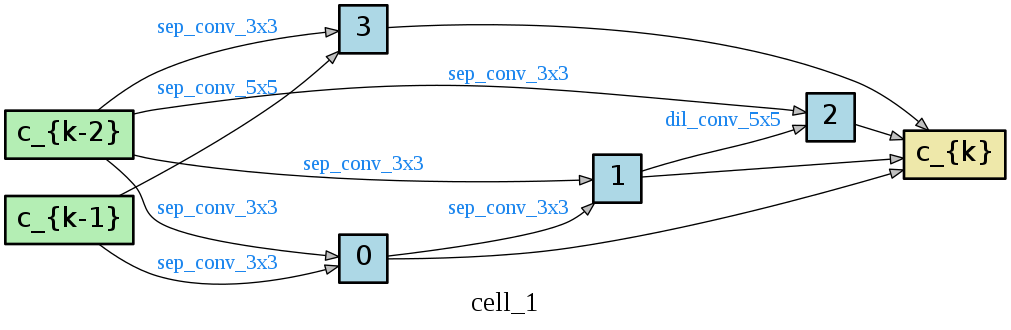} 
       \caption{Reduction cell of CIFAR-10 for S4 (StacNAS) }%
\end{figure}


\begin{figure}[htbp]
\begin{tabular}{cc}
\begin{minipage}[c]{0.5\linewidth}
  \centering
    \includegraphics[width=4cm]{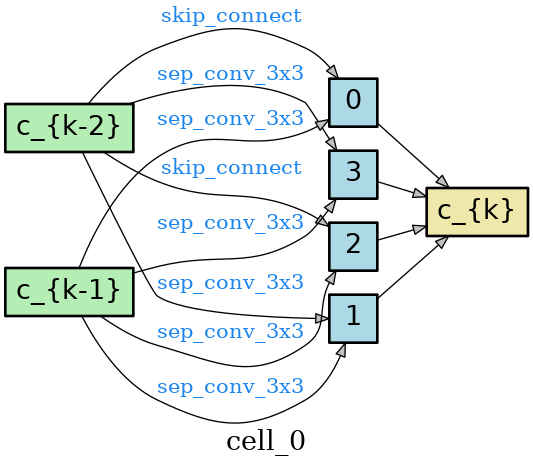} 
       \caption{Normal cell of CIFAR-10 for S5 (StacNAS) }%
     
\end{minipage} 
\begin{minipage}[c]{0.5\linewidth}
  \centering
   \includegraphics[width=6cm]{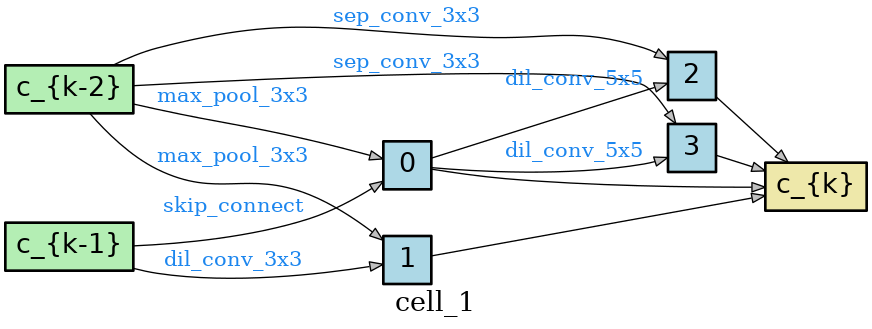} 
       \caption{Reduction cell of CIFAR-10 for S5 (StacNAS) }%
\end{minipage}
\end{tabular}
 
\end{figure}

 \begin{figure}[htbp]
  \centering
   \includegraphics[width=10cm]{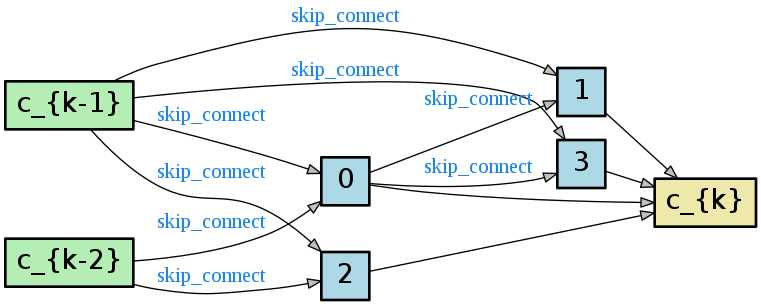} 
       \caption{A cell dominated by SkipConnect discoverd by DARTS }%
\end{figure}
  \begin{figure}[htbp]
  \centering
   \includegraphics[width=10cm]{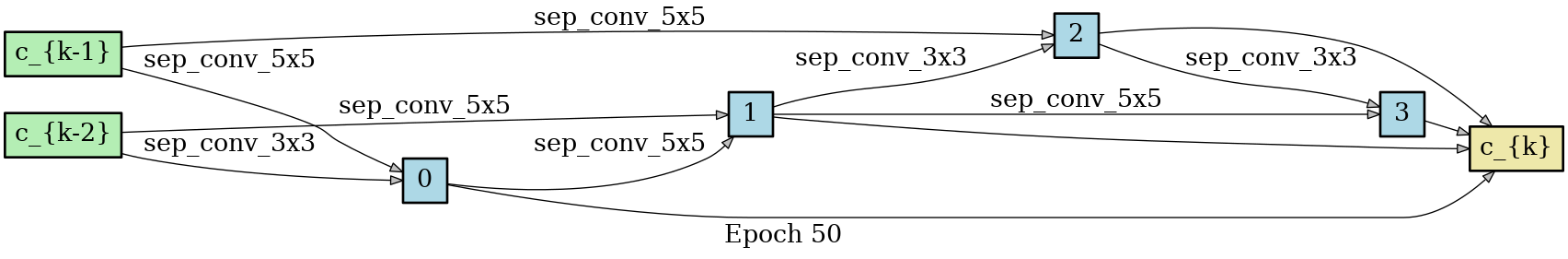} 
       \caption{A cell dominated by SepConv discoverd by DARTS }%
\end{figure}

\end{document}